\documentclass[letterpaper, 10 pt, conference]{ieeeconf}  

\IEEEoverridecommandlockouts                              

\let\labelindent\relax

\usepackage{macros}

\title{\LARGE \bf SOL-SLAM: Inverse Compositional Gauss-Newton Direct Registration for Fast Sonar-Only Local SLAM}

\author{Kalvik Jakkala and Jason O'Kane
\thanks{$^{1}$Kalvik Jakkala and Jason O'Kane are with Texas A\&M University
        {\tt\small \{kalvik, jokane\}@tamu.edu}}%
}

\begin{document}
\maketitle
\thispagestyle{plain}
\pagestyle{plain}

\begin{abstract}
Autonomous underwater navigation typically relies on complex and expensive multi-modal sensor suites designed to prioritize global Simultaneous Localization and Mapping (SLAM) accuracy. However, local reactive behaviors such as coarse navigation and obstacle avoidance require only local consistency—a capability that should be feasible using only a Forward-Looking Sonar (FLS), yet remains largely unaddressed, leaving a critical gap in FLS-only local SLAM.

Moreover, existing acoustic SLAM frameworks predominantly rely on sparse feature extraction methods that discard substantial portions of the already information-sparse acoustic returns. To overcome these limitations, this work introduces a dense direct registration approach that aligns full acoustic intensity scans to a recursively updated local map. Real-time execution is achieved via an Inverse Compositional Gauss-Newton optimization strategy that minimizes compute overhead.

Experimental evaluations show that this dense method yields significant improvements on translation error compared to sparse keypoint baselines, maintaining stable sub-meter tracking precision over wide displacement gaps. Moreover, this approach delivers odometry performance comparable to multi-sensor fusion pipelines (FLS, DVL, and IMU), bypassing expensive payload dependencies in feature-rich environments. We validate real-world applicability through AUV field trials, running the full local SLAM approach onboard an embedded, resource-constrained computer.
\end{abstract}
\section{Introduction}

Reliable environmental perception and localized state estimation are critical for autonomous subsea navigation. However, aquatic mediums rapidly attenuate electromagnetic waves—rendering the Global Positioning System (GPS) and radar infeasible—while turbidity and caustics make optical cameras unreliable. Consequently, Forward-Looking Sonar (FLS) has emerged as the primary perception modality for underwater vehicles~\cite{WangFSNZ23, XuZHLW24}. 

Nonetheless, extracting tracking information from FLS data introduces severe technical challenges: (1) \textbf{Elevation Ambiguity:} The 3D-to-2D projection discards the vertical axis, collapsing geometry and introducing structural ambiguity. (2) \textbf{Speckle Noise:} Coherent wave reflections generate high-frequency interference that corrupts structural boundaries. (3) \textbf{Acoustic Shadows:} Geometric occlusions cast viewpoint-dependent, zero-signal shadows that shift dynamically with motion. (4) \textbf{Non-Uniform Resolution:} Azimuthal resolution degrades linearly with range due to beam fanning, dilating far-field returns.

To mitigate these factors, contemporary underwater Simultaneous Localization and Mapping (SLAM) systems~\cite{RahmanLR19, WangCHMSE22, XuZHLW24} rely on complex, cost-prohibitive sensor suites fusing cameras, sonars, Inertial Measurement Units (IMUs), Doppler Velocity Logs (DVLs), Ultra-Short Baseline (USBL) systems, and depth sensors to compensate for individual sensor limitations and ensure global accuracy. However, local reactive behaviors like obstacle avoidance require only local accuracy—a capability that should be viable using only an FLS, without the cost and complexity of multi-sensor fusion. Consequently, a distinct research gap exists for a lightweight, FLS-based local SLAM approach.

This work introduces a local SLAM approach operating exclusively on FLS data to yield real-time relative motion estimates for standalone coarse navigation and obstacle avoidance. To overcome sparse keypoint extractors (e.g., AKAZE~\cite{AlcantarillaNB13} or SONIC~\cite{GodeHK24}), which discard valuable geometric context, we leverage a dense direct registration approach that aligns full acoustic intensity scans to a recursively updated local map. To achieve real-time execution, we formulate the registration via an Inverse Compositional Gauss-Newton optimization strategy~\cite{BakerM04}. This shifts spatial gradient calculations onto the static map, enabling fast second-order convergence. Crucially, by actively updating this dense local map with statistical noise decay, our approach goes beyond simple frame-to-frame odometry to directly provide the spatial awareness required for closed-loop navigation.

\begin{figure}[!t]
\centering
\includegraphics[width=\linewidth]{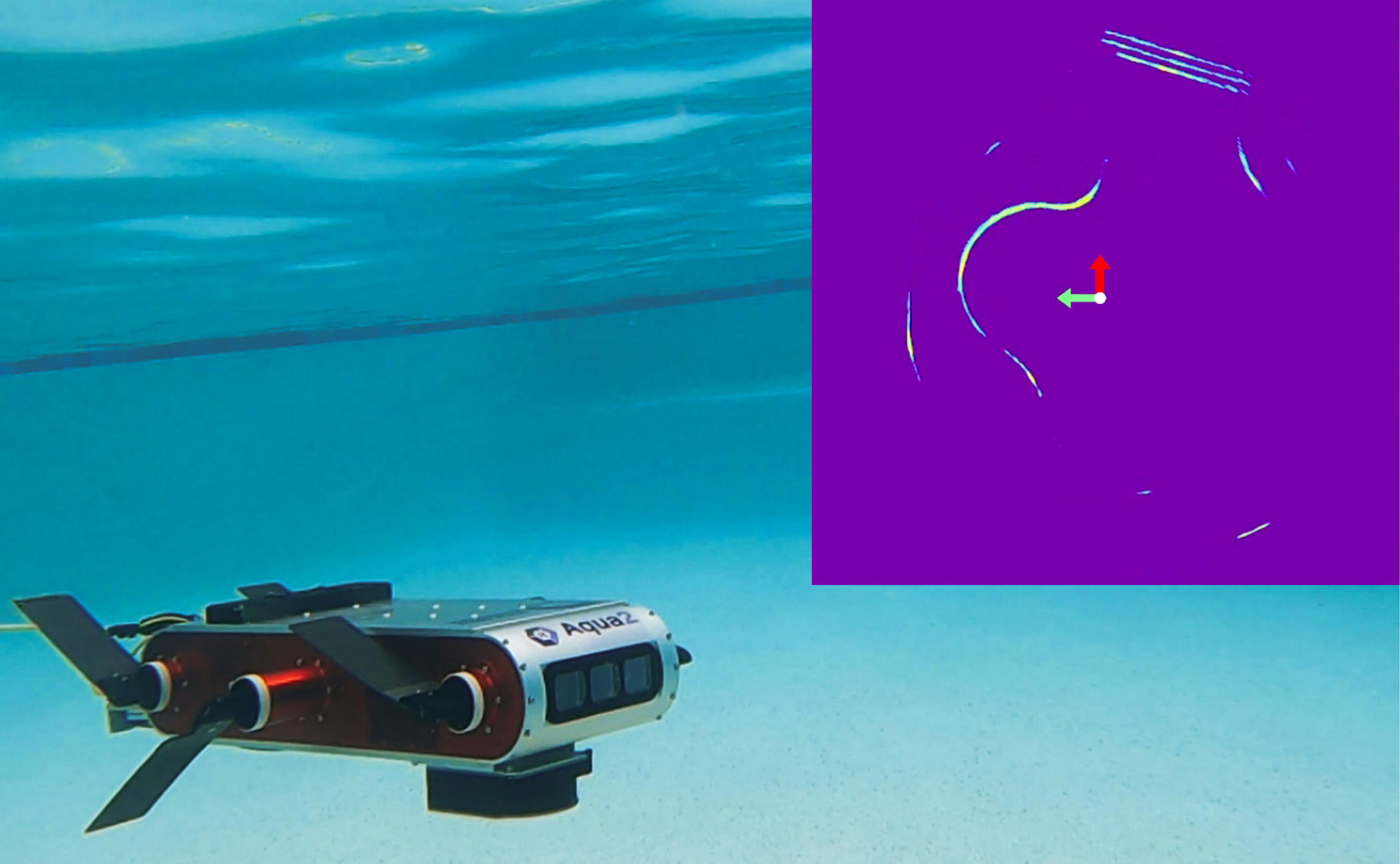}
\caption{Aqua2 AUV during the field trial (left) alongside the local map estimated by SOL-SLAM (right).}
\label{fig:aqua}
\vspace{-4mm}
\end{figure}

The primary contributions of this work are as follows:
\begin{itemize}
    \item \textbf{SOL-SLAM (Sonar-Only Local SLAM):} We present the first FLS-only, direct registration-based local SLAM approach, which formulates scan matching as cross-correlation and solves it using an efficient inverse-compositional Gauss–Newton algorithm. By operating directly on full sonar scans rather than sparse features, our approach achieves robust real-time performance in feature-degraded environments.
    
    \item \textbf{Field Trial Validation:} We validate the feasibility of our approach through a real-world deployment on an Autonomous Underwater Vehicle (AUV), relying solely on an FLS and onboard compute.
\end{itemize}
\section{Related Work}

Constrained by GPS denial and acoustic noise, subsea navigation typically utilizes tightly coupled multi-modal architectures~\cite{WangFSNZ23, HeshmatSASESH25}. This section reviews key methods, contrasting them with our SOL-SLAM approach.

\subsection{Odometry Tracking vs. SLAM Architecture}

\begin{itemize}[leftmargin=*, nosep, topsep=0pt, partopsep=0pt]
    \item \textbf{Odometry Tracking:} Subsea odometry methods generally leverage multi-modal sensor fusion to bound local tracking drift. For example, DISO~\cite{XuZHLW24} minimizes sonar scan intensity gradients by incorporating synchronized Doppler Velocity Log (DVL) and Inertial Measurement Unit (IMU) inputs.
    
    \item \textbf{SLAM Architectures:} Simultaneous Localization and Mapping (SLAM) systems maintain global consistency across extended vehicle trajectories. These architectures typically utilize incremental smoothing techniques like iSAM2~\cite{HeshmatSASESH25}, Expectation-Maximization loops (EM-loops) with virtual landmarks~\cite{WangCHMSE22}, or tightly coupled multi-sensor optimizations~\cite{RahmanLR19, yangHLG26}, which jointly processes visual, inertial, sonar, and depth data.
\end{itemize}
\vspace{1mm}
\noindent\textbf{Distinction:} Existing approaches rely on complex multi-modal sensor suites for tracking stability~\cite{XuZHLW24, BaiZCL25, RahmanLR19, HeshmatSASESH25} or incur considerable latency via global factor graph optimization~\cite{HeshmatSASESH25} to ensure global accuracy. In contrast, the proposed SOL-SLAM approach focuses on locally accurate SLAM, which allows it to omit global factorizations and rely only on a single Forward-Looking Sonar (FLS), delivering real-time spatial awareness for local navigation.

\subsection{Feature Extraction}

\begin{itemize}[leftmargin=*, nosep, topsep=0pt, partopsep=0pt]
    \item \textbf{Handcrafted Primitives:} Traditional handcrafted methods isolate distinct structural transitions, such as acoustic edge-shadow boundaries \cite{JohannssonKEHL10}, or employ Contrast-Limited Adaptive Histogram Equalization to enhance visual keypoints \cite{RahmanLR19}.
    \item \textbf{Direct Density Fields:} Direct methods minimize intensity errors across dense pixel grids. These approaches include extracting raw spatial gradients \cite{XuZHLW24}, tracking whole-frame intensity displacement fields via optical flow \cite{HensonZ19}, or registering raw intensity scans directly in the context of radar data \cite{GentilBLQGB25}.
    \item \textbf{Learned Operators:} Deep neural networks extract data-driven feature representations by utilizing specialized polar range-bearing image descriptors \cite{GodeHK24}, cross-branch attention mechanisms \cite{YaoLSWY26}, or Multi-scale Constant False Alarm Rate (MCFAR) filters \cite{BaiZCL25}.
\end{itemize}
\vspace{1mm}
\noindent\textbf{Distinction:} Sparse handcrafted and learned descriptors are highly vulnerable to underwater speckle noise and tracking dropouts \cite{YaoLSWY26, XuZHLW24, GodeHK24}. While direct methods like DRO were explicitly meant for radar modalities, our approach shows how to effectively generalize these direct, featureless representations to FLS sonar data.

\subsection{Scan Registration}

\begin{itemize}[leftmargin=*, nosep, topsep=0pt, partopsep=0pt]
    \item \textbf{Geometric Primitives:} Traditional geometric methods align structural geometry by applying the Iterative Closest Point (ICP) algorithm to associate and align extracted contours, utilizing the Normal Distributions Transform (NDT) across discrete grids \cite{JohannssonKEHL10}, or projecting 2D FLS scans into 3D point clouds to perform joint point-to-line or point-to-surface matching \cite{BaiZCL25}. To mitigate drift across multiple scans, some geometric pipelines additionally incorporate local bundle adjustment to jointly optimize consecutive sensor poses and mapped spatial landmarks over a sliding window~\cite{BaiZCL25}.
    \item \textbf{Direct Photometric Alignment:} Photometric techniques optimize relative transformations directly on the motion manifold using dead-reckoning priors \cite{XuZHLW24}, perform scan-to-local-map registration directly on radar intensity information \cite{GentilBLQGB25}, or employ coarse pixel-level Orthogonal Matching Pursuit coupled with Recursive Least Squares adaptive filtering \cite{HensonZ19}.
    \item \textbf{Frequency-Domain Alignment:} Spectral methods resolve relative pose via phase correlation \cite{HurtosCPS12} or project spectral magnitudes into the Radon domain using learnable upsampling and phase estimation modules \cite{YaoLSWY26}.
\end{itemize}
\vspace{1mm}
\noindent\textbf{Distinction:} While geometric registration is highly sensitive to initialization \cite{JohannssonKEHL10, BaiZCL25} and spectral methods require computationally demanding coordinate resampling \cite{HensonZ19, HurtosCPS12, YaoLSWY26}, the proposed SOL-SLAM approach mitigates both issues. It avoids the heavy computational costs of generalized direct methods through an Inverse Compositional Gauss-Newton formulation, which enables the precomputation of spatial gradients on a smoothed local map. Coupled with Gauss-Newton updates, this precomputation drastically accelerates inference, efficiently resolving motion parameters over dense grids for sub-pixel precision and minimal overhead.

\subsection{Loop Closure}

\begin{itemize}[leftmargin=*, nosep, topsep=0pt, partopsep=0pt]
    \item \textbf{Handcrafted Vocabularies and Gated Matrices:} 
    
    Some methods compress descriptive features into searchable indices by querying a binary Bag-of-Words (DBoW2) database~\cite{RahmanLR19}, while others gate loop candidates via Peak-to-Noise Ratio (PNR) thresholds derived from spectral correlation~\cite{HurtosCPS12}.
    
    \item \textbf{Learned and Pose-Supervised Tracking:} Recent techniques automate place recognition using convolutional Siamese networks~\cite{HeshmatSASESH25}, Variational Autoencoders (VAEs) with semantic dynamic filtering~\cite{HeshmatSASESH25}, or pose-supervised geometric objectives—such as SONIC~\cite{GodeHK24}—that minimize epipolar distance loss.
\end{itemize}

\begin{figure*}[!ht]
\centering
\begin{subfigure}{0.49\linewidth}
\includegraphics[width=\textwidth]{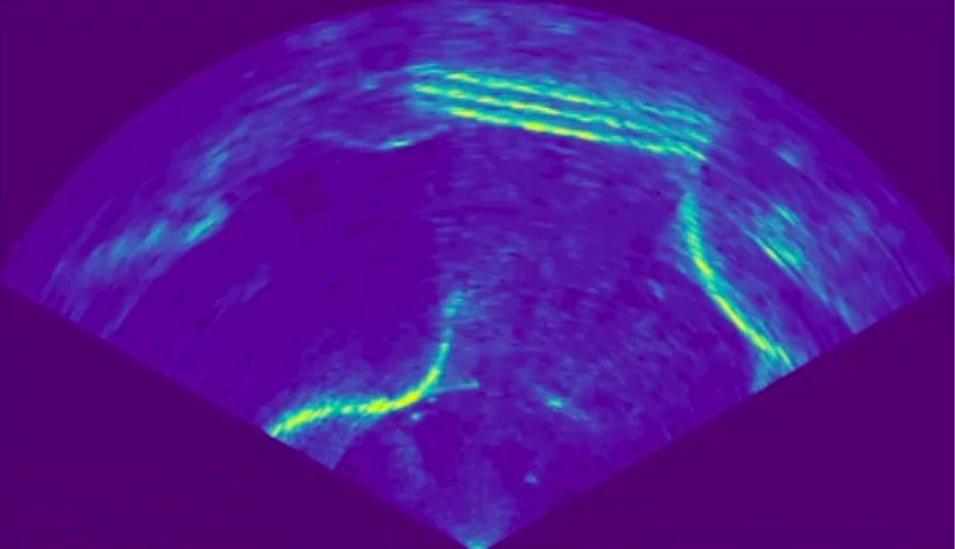}
\end{subfigure}
\hfill
\begin{subfigure}{0.49\linewidth}
\includegraphics[width=\textwidth]{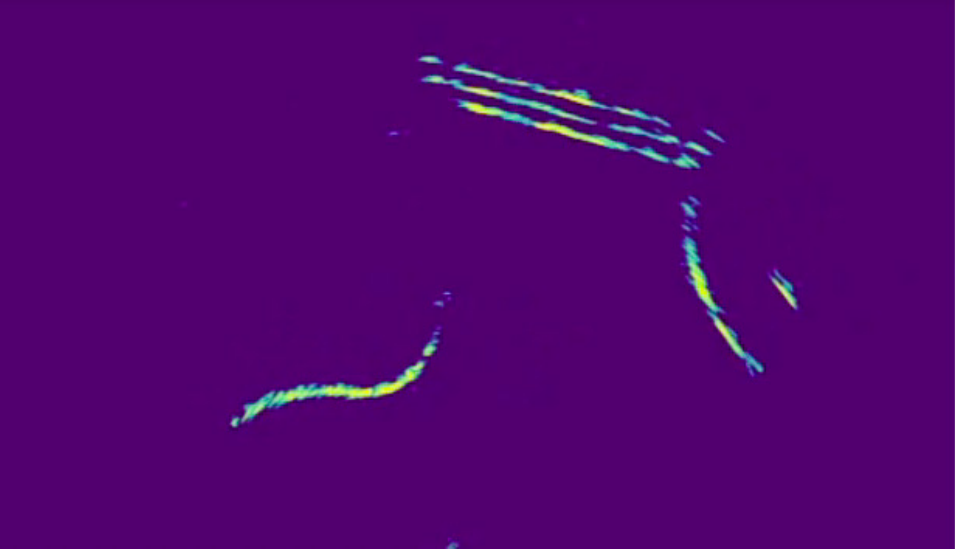}
\end{subfigure}
\caption{Forward-Looking Sonar (FLS) scans: raw (left) and denoised (right).}\label{fig:sonar_scans}
\end{figure*}

\vspace{1mm}
\noindent\textbf{Distinction:} Global loop closure introduces critical processing bottlenecks due to continuous dictionary queries, deep neural network inference, and batch factor graph optimizations, presenting a challenge for real-time deployment on embedded subsea computing stacks \cite{HeshmatSASESH25}. The proposed SOL-SLAM approach intentionally omits global loop closure and global pose-graph relaxation to minimize computational overhead. By registering dense scans directly against a rolling local map, our approach bounds tracking drift locally to realize a fast, locally consistent SLAM approach optimized for real-time navigation and reactive hazard avoidance. 

For broader surveys of underwater mapping and estimation paradigms, see \cite{WangFSNZ23, HeshmatSASESH25, Santos25}.
\section{Background}
\label{sec:background}

\subsection{FLS Operation and Polar Representation}

A Forward-Looking Sonar (FLS) scans subsea environments by emitting acoustic pings across a horizontal field of view $\Phi_{\text{fov}}$ with a narrow vertical beamwidth ($1^\circ\text{--}3^\circ$). This process yields a complete acoustic scan represented as a 2D polar matrix $P \in [0, 1]^{N_r \times N_\phi}$, where $N_r$ and $N_\phi$ denote the total number of range bins and discrete acoustic beams, respectively. Individual bin intensities $P_{i,j}$ map directly to backscatter amplitudes and are indexed by discrete range $r_i$ and bearing $\phi_j$:

$$r_i \in [R_{\min}, R_{\max}], \quad \phi_j \in \left[-\frac{\Phi_{\text{fov}}}{2}, \frac{\Phi_{\text{fov}}}{2}\right]\,.$$

\subsection{Cartesian Space Mapping}

To perform state estimation, the polar scan $P$ is mapped into a dense Cartesian image $I$. A continuous polar coordinate pair $(r, \phi)$ maps to a spatial coordinate vector $\mathbf{u} = [x, y]^T$ via forward geometric projection:

$$x = r \cos\phi, \quad y = r \sin\phi\,.$$

To avoid the non-uniform spatial sampling inherent to direct forward projection, a backward-lookup mapping is used to construct a uniformly discretized image $I$:

$$r = \sqrt{x^2 + y^2}, \quad \phi = \operatorname{arctan2}(y, x)\,.$$

The final intensity value for an individual pixel $I(\mathbf{u})$ is computed via bilinear interpolation over the four surrounding discrete bins $P_{i,j}$. This transformation yields a spatially consistent dense image $I$ optimized for direct, intensity-based tracking.

\subsection{Kinematic Warp Function}

Let $M: \Omega \to [0, 1]$ denote the local map defined over a bounded Cartesian spatial domain $\Omega = [X_{\min}, X_{\max}] \times [Y_{\min}, Y_{\max}] \subset \mathbb{R}^2$, where $M(\mathbf{u})$ assigns an acoustic intensity to each coordinate vector $\mathbf{u} = [x, y]^T \in \Omega$. Let $\mathbf{x} = [t_x, t_y, \theta]^T \in \mathbb{R}^2 \times \mathbb{S}^1$ define the state vector parameterizing the vehicle pose in $\mathrm{SE}(2)$ relative to the local map frame, where $\mathbb{S}^1$ denotes the angular domain. The kinematic warp function $W(\mathbf{u}; \mathbf{x})$ defines a rigid 2D transformation mapping a spatial coordinate vector $\mathbf{u} \in \Omega$ from the local map frame into the FLS sensor frame:

$$W(\mathbf{u}; \mathbf{x}) = \begin{bmatrix} \cos\theta & -\sin\theta \\ \sin\theta & \cos\theta \end{bmatrix} \mathbf{u} + \begin{bmatrix} t_x \\ t_y \end{bmatrix}\,.$$
\section{SOL-SLAM (Sonar-Only Local SLAM)}

This section presents our Forward-Looking Sonar (FLS)-only local SLAM approach. Using streaming sonar scans from an FLS without auxiliary sensing, the approach concurrently maintains a robot-centric local map and tracks the vehicle's relative $\mathrm{SE}(2)$ pose. To ensure real-time execution for reactive obstacle avoidance, we formulate dense scan-to-map alignment as a cross-correlation optimization problem, solved efficiently using an Inverse Compositional Gauss–Newton (IC-GN) formulation.

\subsection{Data Preprocessing and Dense Feature Extraction}
Raw FLS scans are highly susceptible to acoustic speckle noise and coherent phase interference. To prevent these signal-level artifacts from corrupting direct registration, a background suppression filter is applied directly to the polar matrix prior to coordinate transformation:
\begin{equation}
    P^{\text{clean}}_{i,j} = \begin{cases} P_{i,j} & \text{if } P_{i,j} \ge \mu_P + \beta_{\text{denoise}} \sigma_P, \\ 0 & \text{otherwise}, \end{cases}
\end{equation}
where $P_{i,j}$ represents the raw polar intensity at range $r_i$ and bearing $\phi_j$, while $\mu_P$ and $\sigma_P$ denote the mean and standard deviation of the polar scan intensity distribution, respectively. The scalar $\beta_{\text{denoise}}$ represents an empirical structural denoising coefficient.

Following polar scan filtering, the denoised matrix $P^{\text{clean}}$ is mapped to Cartesian space via bilinear interpolation, yielding a dense intensity image $I(\mathbf{u})$ structured for direct registration (Figure~\ref{fig:sonar_scans}).

\subsection{Scan-to-Map Alignment Optimization Bottleneck}
At the start of a mapping sequence, the reference map $M(\mathbf{u})$ is initialized directly from the first preprocessed Cartesian scan. Subsequent incoming scans $I(\mathbf{u})$ are then aligned against this reference map. Existing direct alignment approaches establish scan-to-map registration by optimizing the vehicle pose vector $\mathbf{x} \in \mathbb{R}^2 \times \mathbb{S}^1$ via a standard $L_2$ photometric loss:
\begin{equation}
    \mathcal{L}_{L_2}(\mathbf{x}) = \frac{1}{2} \sum_{\mathbf{u} \in \Omega} \left( I\big(W(\mathbf{u}; \mathbf{x})\big) - M(\mathbf{u}) \right)^2,
\end{equation}
where $W(\mathbf{u}; \mathbf{x})$ is the rigid 2D kinematic warp function mapping coordinates from the local map frame to the sensor frame, as detailed in Section~\ref{sec:background}.

However, the $L_2$ norm exhibits poor robustness when applied to FLS imagery. Acoustic shadows cast by topography or physical hazards yield zero-intensity regions that shift dynamically with the vehicle's changing perspective. A standard difference norm penalizes these unsonified zones, introducing artificial local minima that degrade tracking accuracy. To overcome this limitation, we formulate the registration objective using Forward Compositional (FC) cross-correlation over the spatial domain, parameterized by an incremental pose adjustment $\Delta \mathbf{x}$ defined in the map frame:
\begin{equation}
    \mathcal{L}_{\text{FC}}(\Delta \mathbf{x}) = -\frac{1}{|\Omega|} \sum_{\mathbf{u} \in \Omega} I\big(W(W(\mathbf{u}; \Delta \mathbf{x}); \mathbf{x})\big) M(\mathbf{u}).
\end{equation}

Here, the nested warp $W(W(\mathbf{u}; \Delta \mathbf{x}); \mathbf{x})$ applies an incremental perturbation $\Delta \mathbf{x}$ to the map coordinate $\mathbf{u}$ prior to projecting it into the sensor frame via the current pose estimate $\mathbf{x}$. Evaluating this intensity inner product ensures that zero-signal shadow zones drop out of the summation naturally ($M(\mathbf{u}) = 0$ or $I(\cdot) = 0$) rather than biasing the residual, rendering the tracking approach robust to fluctuating acoustic illumination and non-overlapping shadows.

To optimize this objective, evaluating the first-order derivative of the forward compositional loss $\mathcal{L}_{\text{FC}}$ with respect to $\Delta \mathbf{x}$ at $\Delta \mathbf{x} = \mathbf{0}$ yields the gradient vector:
\begin{equation}
    \mathbf{g}_{\text{FC}} = \left. \frac{\partial \mathcal{L}_{\text{FC}}}{\partial \Delta \mathbf{x}} \right|_{\Delta \mathbf{x}=\mathbf{0}} = -\frac{1}{|\Omega|} \sum_{\mathbf{u} \in \Omega} M(\mathbf{u}) \mathbf{J}_{\text{FC}}(\mathbf{u}; \mathbf{x}),
\end{equation}
where $\mathbf{J}_{\text{FC}}(\mathbf{u}; \mathbf{x}) \in \mathbb{R}^{3 \times 1}$ represents the dynamic forward compositional steepest descent vector:
\begin{equation*}
    \mathbf{J}_{\text{FC}}(\mathbf{u}; \mathbf{x}) = \left( \nabla I\big(W(\mathbf{u}; \mathbf{x})\big) \begin{bmatrix} \cos\theta & -\sin\theta \\ \sin\theta & \cos\theta \end{bmatrix} \begin{bmatrix} 1 & 0 & -y \\ 0 & 1 & x \end{bmatrix} \right)^T
\end{equation*}

An inspection of $\mathbf{J}_{\text{FC}}(\mathbf{u}; \mathbf{x})$ reveals a severe computational bottleneck. Because $\Delta \mathbf{x}$ represents only a local step evaluated around the current linearization point, the tracking pose must be updated at each iteration ($\mathbf{x} \leftarrow \mathbf{x} \circ \Delta \mathbf{x}$), shifting the warped coordinates $W(\mathbf{u}; \mathbf{x})$. Although the raw scan $I$ is static, updating $\mathbf{x}$ shifts these projection points to new continuous, sub-pixel locations. Consequently, the spatial gradients $\nabla I(W(\mathbf{u}; \mathbf{x}))$ must be repeatedly interpolated across all pixels, forcing the steepest descent vector to be re-computed. This per-iteration recalculation introduces severe computational overhead.

\subsection{Inverse Compositional Registration Approach}
We eliminate the computational bottleneck by executing a change of variables under an Inverse Compositional (IC) approach. An alternative coordinate space is defined to absorb the incremental parameter perturbation:
\begin{equation}
    \mathbf{u}' = W(\mathbf{u}; \Delta \mathbf{x}) \implies \mathbf{u} = W(\mathbf{u}'; \Delta \mathbf{x})^{-1}.
\end{equation}
Because $W$ represents a strictly rigid 2D transformation, it defines a bijective mapping over the spatial grid $\Omega$, preserving the domain area. Substituting this coordinate transformation into the objective function and renaming the index variable of summation back to $\mathbf{u}$ yields the IC registration objective:
\begin{equation}
    \mathcal{L}_{\text{IC}}(\Delta \mathbf{x}) = -\frac{1}{|\Omega|} \sum_{\mathbf{u} \in \Omega} I\big(W(\mathbf{u}; \mathbf{x})\big) M\big(W(\mathbf{u}; \Delta \mathbf{x})\big).
\end{equation}
Through this mathematical substitution, the incremental parameter updates are shifted from the dynamic scan $I$ onto the static reference map $M$. Linearizing the map intensity expression around an identity perturbation ($\Delta \mathbf{x} = \mathbf{0}$) via a first-order Taylor expansion yields:
\begin{equation}
    M\big(W(\mathbf{u}; \Delta \mathbf{x})\big) \approx M(\mathbf{u}) + \nabla M(\mathbf{u}) \frac{\partial W(\mathbf{u}; \Delta \mathbf{x})}{\partial \Delta \mathbf{x}} \Delta \mathbf{x},
\end{equation}
where $\nabla M(\mathbf{u}) = [M_x(\mathbf{u}), M_y(\mathbf{u})]$ represents the spatial gradient vector of the local map. Projecting these spatial derivatives onto the rigid 2D geometric Jacobian yields the constant steepest descent image vector $\mathbf{J}(\mathbf{u}) \in \mathbb{R}^{3 \times 1}$:
\begin{equation*}
    \mathbf{J}(\mathbf{u}) = \left( \nabla M(\mathbf{u}) \begin{bmatrix} 1 & 0 & -y \\ 0 & 1 & x \end{bmatrix} \right)^T = \begin{bmatrix} M_x(\mathbf{u}) \\ M_y(\mathbf{u}) \\ -y M_x(\mathbf{u}) + x M_y(\mathbf{u}) \end{bmatrix}
\end{equation*}
The first-order gradient of the alignment loss, $\mathbf{g} \in \mathbb{R}^{3 \times 1}$, evaluated at $\Delta \mathbf{x} = \mathbf{0}$, is derived directly as:
\begin{equation}
    \mathbf{g} = \left. \frac{\partial \mathcal{L}_{\text{IC}}}{\partial \Delta \mathbf{x}} \right|_{\Delta \mathbf{x}=\mathbf{0}} = -\frac{1}{|\Omega|} \sum_{\mathbf{u} \in \Omega} I\big(W(\mathbf{u}; \mathbf{x})\big) \mathbf{J}(\mathbf{u}).
\end{equation}
Crucially, because the local map $M$ remains static during the registration of a single incoming scan, both the spatial map gradients and the steepest descent vectors $\mathbf{J}(\mathbf{u})$ are constant. They are evaluated exactly once per incoming scan, unlocking fast optimization speeds at a fraction of the traditional computational cost.

\subsection{Gauss-Newton Optimization Loop}
To enable fast alignment across dense spaces, we employ a Gauss-Newton optimization approach. While traditional direct alignment methods evaluate a dynamic Hessian at every optimization iteration, the Inverse Compositional (IC) formulation enables the Hessian matrix, $\mathbf{H} \in \mathbb{R}^{3 \times 3}$, to be approximated using the outer product of the precomputed first-order steepest descent vectors, bypassing expensive second-order derivative computations:
\begin{equation}
    \mathbf{H} \approx \frac{1}{|\Omega|} \sum_{\mathbf{u} \in \Omega} \mathbf{J}(\mathbf{u}) \mathbf{J}(\mathbf{u})^T.
\end{equation}
To prevent matrix singularities and numerical instability when operating over feature-sparse or geometrically degenerate topologies, Levenberg-Marquardt (LM) damping is applied via a regularization factor $\lambda$:
\begin{equation}
    \mathbf{H}_{\text{LM}} = \mathbf{H} + \lambda \mathbb{I},
\end{equation}
where $\mathbb{I}$ denotes the $3 \times 3$ identity matrix. Because $\mathbf{H}$ and its regularized inverse $\mathbf{H}_{\text{LM}}^{-1}$ depend solely on the static map parameters, they are computed and inverted exactly once per incoming scan prior to entering the iterative tracking loop. Inside the loop, the pose state vector is updated iteratively at iteration $i$ via:
\begin{equation}
    \mathbf{x}^{(i+1)} = \mathbf{x}^{(i)} - \gamma \mathbf{H}_{\text{LM}}^{-1} \mathbf{g}^{(i)},
\end{equation}
where $\gamma$ represents the operational step size.

Acoustic speckle noise frequently introduces random intensity spikes that can inject destabilizing outliers into the image gradients. We counter this issue by implementing a dynamic gradient clipping scheme. If the Euclidean norm of the first-order gradient exceeds a maximum allowable threshold $g_{\max}$, the vector is scaled to preserve its directional trajectory while enforcing numerical stability:
\begin{equation}
    \mathbf{g}^{(i)} \leftarrow \mathbf{g}^{(i)} \min\left(1, \, \frac{g_{\max}}{\|\mathbf{g}^{(i)}\|}\right).
\end{equation}
The internal optimization loop terminates when the parameter update magnitude $\|\mathbf{x}^{(i+1)} - \mathbf{x}^{(i)}\|$ drops below a specified tolerance limit or when the fractional change in the correlation loss satisfies $\Delta \mathcal{L}_{\text{IC}} < \epsilon_{\text{tol}}$.

\subsection{Kinematic State Propagation and Map Maintenance}
Following the convergence of the optimization approach, the vehicle's relative displacement over the tracking interval is processed by a kinematic motion model to maintain temporal trajectory consistency across sequential scans. Assuming a constant velocity profile over the inter-scan interval $\Delta t$, the local-frame displacement vector $\mathbf{x}_k = [t_x, t_y, \theta]^T$ is converted into linear and angular velocities:
\begin{equation}
    v_x = \frac{t_x}{\Delta t}, \quad v_y = \frac{t_y}{\Delta t}, \quad v_\theta = \frac{\theta}{\Delta t}.
\end{equation}

These velocity components update the trajectory estimate and provide a predictive warm-start initialization $\mathbf{x}_{k+1}^{(0)}$ for the subsequent registration cycle, effectively seeding the optimization within its basin of attraction:
\begin{equation}
    \mathbf{x}_{k+1}^{(0)} = \Delta t \begin{bmatrix} v_x & v_y & v_\theta \end{bmatrix}^T.
\end{equation}

Following the state propagation step, map maintenance is performed by projectively warping the current preprocessed scan into the map frame, $I_k^{\text{aligned}}(\mathbf{u}) = I_k\big(W(\mathbf{u}; \mathbf{x}_k)\big)$, and recursively integrating it into the local map via an Exponential Moving Average (EMA):
\begin{equation}
    M_k(\mathbf{u}) = (1 - \alpha) M_{k-1}(\mathbf{u}) + \alpha I_k^{\text{aligned}}(\mathbf{u}),
\end{equation}
where $\alpha \in (0, 1]$ represents the map update coefficient.

Because acoustic speckle noise accumulates over extended temporal horizons, direct integration can saturate the local map with high-frequency noise and transient landmarks. To counter this saturation and preserve sharp geometric hazard boundaries, a statistical noise decay filter is applied immediately following the EMA update to dynamically prune intensities based on global grid metrics:
\begin{equation}
    M_k(\mathbf{u}) \leftarrow \begin{cases} M_k(\mathbf{u}) & \text{if } M_k(\mathbf{u}) \ge \mu_M + \beta_{\text{decay}} \sigma_M, \\ 0 & \text{otherwise}, \end{cases}
\end{equation}
where $\mu_M$ and $\sigma_M$ denote the mean and standard deviation of the local map intensity field, respectively, and $\beta_{\text{decay}}$ represents the empirical decay coefficient.

Finally, the reference map is transformed to the updated vehicle pose via bilinear interpolation, and an obstacle point cloud is extracted by thresholding high-intensity acoustic returns into 2D points.
\section{Benchmarks and Field Trial}

This section presents the experimental validation of SOL-SLAM. Since our approach addresses local rather than global SLAM, we evaluate performance over trajectory sub-sequences that do not necessitate loop closure.

\begin{figure*}[!ht]
\centering
\includegraphics[width=\linewidth]{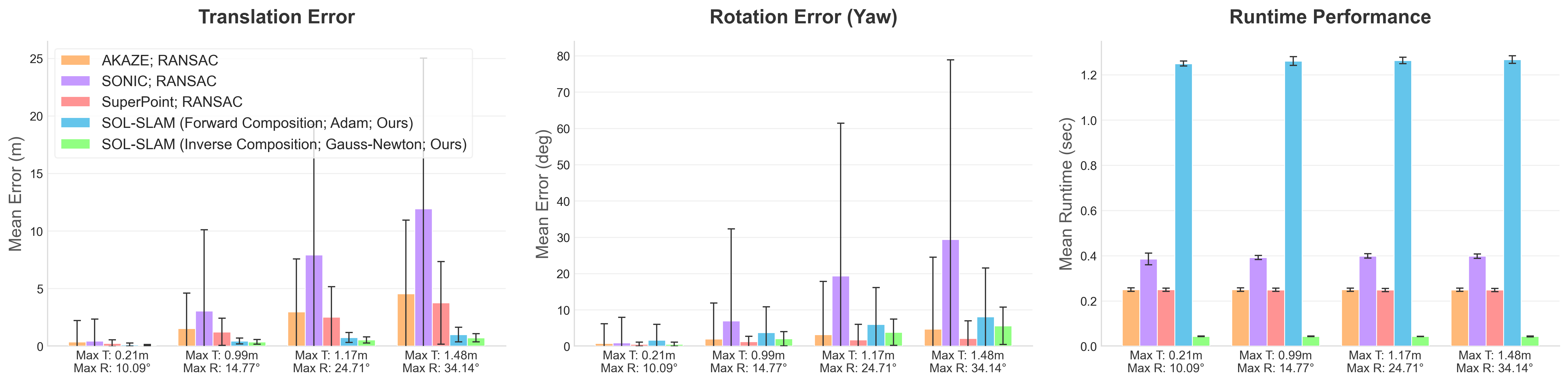}
\caption{Quantitative evaluation under progressive initial misalignments across 5-, 10-, and 15-frame separation gaps, bounded by maximum translation $T$ and rotation $R$. Plots compare mean translation error, mean rotation error, and average execution runtime. The results highlight that our dense SOL-SLAM approach considerably minimizes translation error relative to all evaluated baselines, maintaining stable tracking precision where sparse methods degrade.}
\label{fig:aracati_bar}
\end{figure*}

\begin{figure*}[ht]
\centering
\begin{subfigure}{0.49\linewidth}
\includegraphics[width=\textwidth]{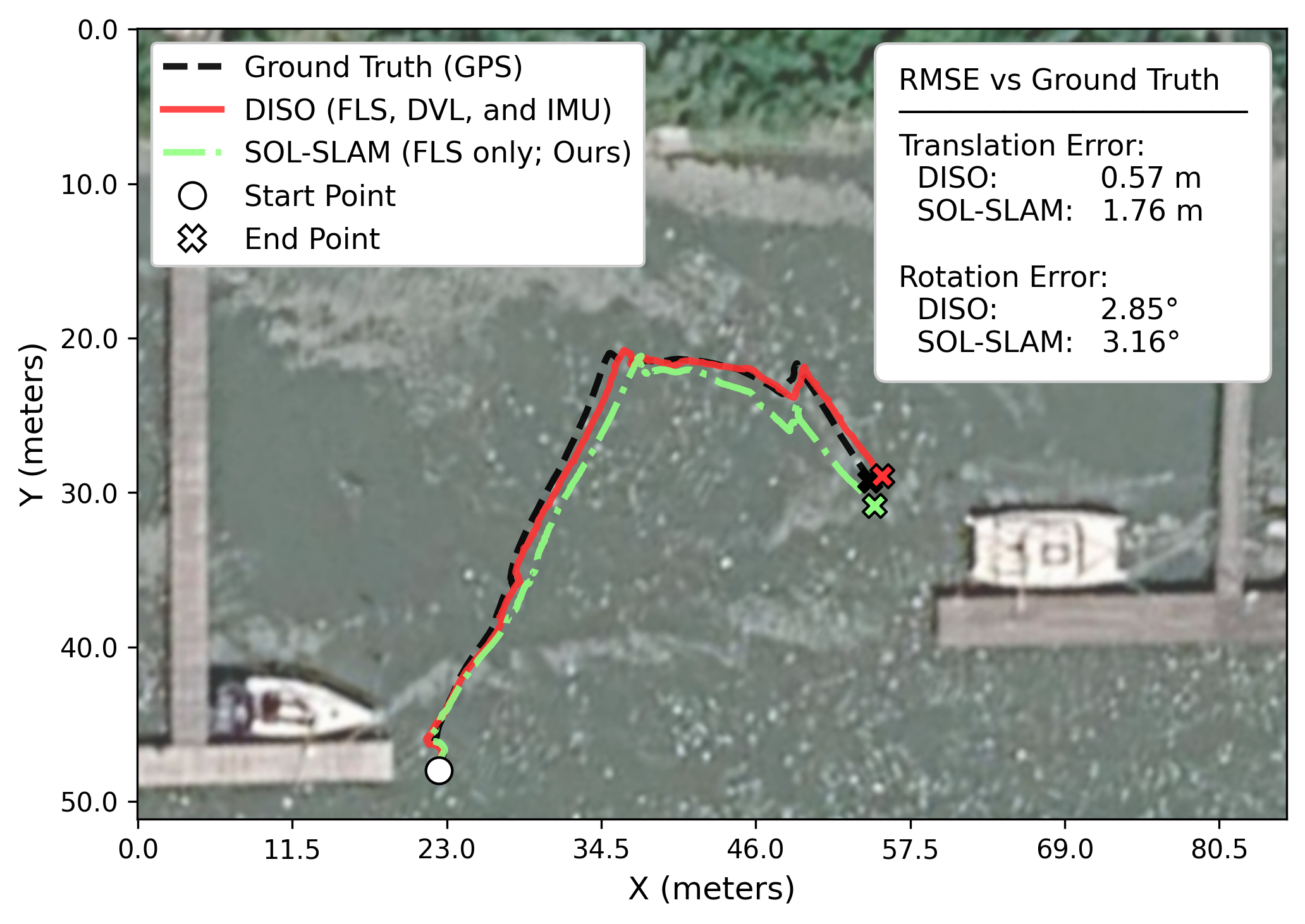}
\end{subfigure}
\hfill
\begin{subfigure}{0.49\linewidth}
\includegraphics[width=\textwidth]{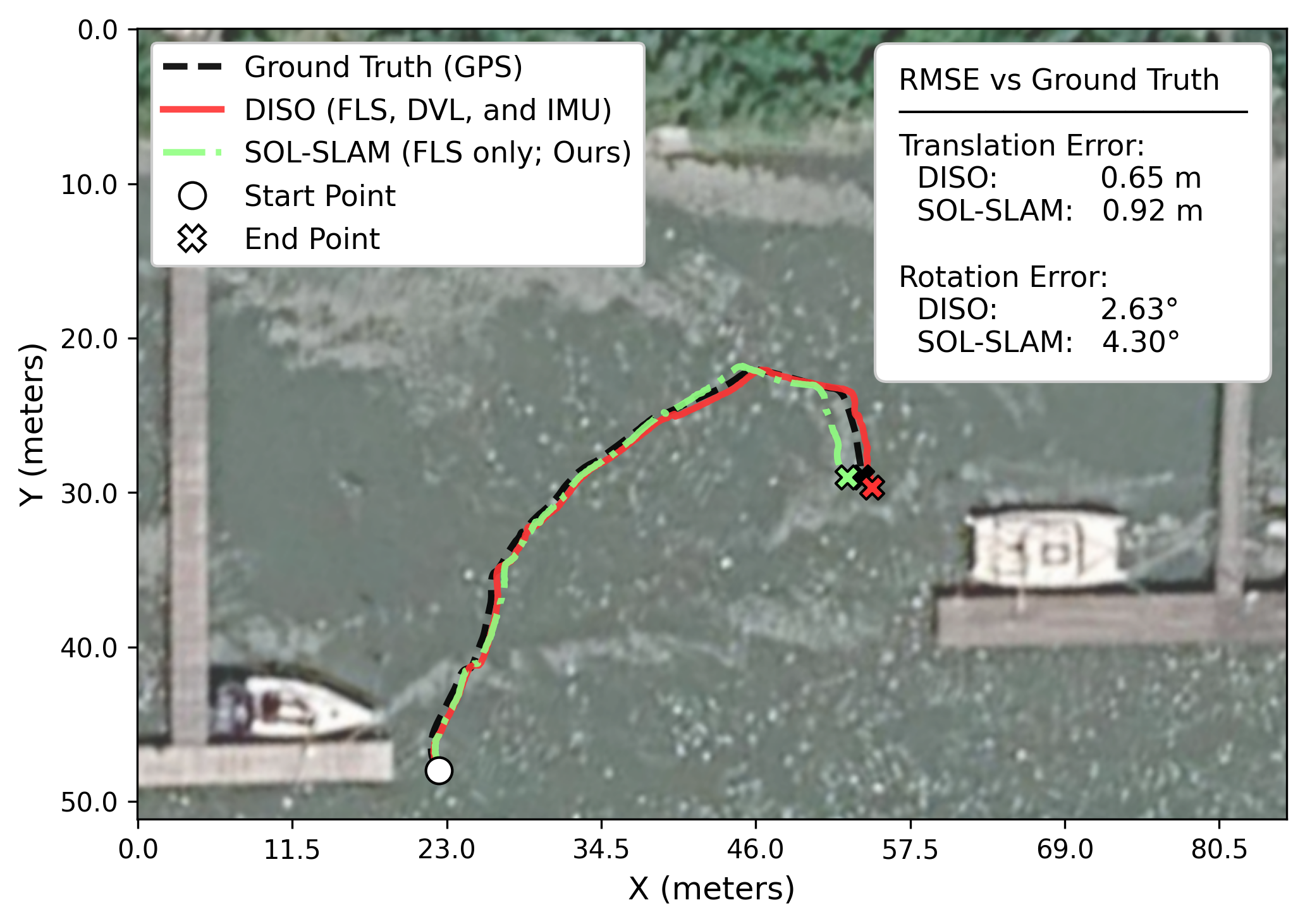}
\end{subfigure}
\caption{Estimated local trajectory traces mapped against Ground Truth (GPS) across two distinct sub-sequences from the Aracati dataset. The metrics demonstrate that our FLS-only local SLAM performance remains competitive with the multi-modal SLAM (DISO).}
\label{fig:aracati_trace}
\vspace{-3mm}
\end{figure*}

\subsection{Benchmark Setup}

The proposed SOL-SLAM approach is evaluated using the Aracati dataset~\cite{ChameSB18}, which was collected at the Yacht Club marina in Rio Grande, Brazil, 2017. The experimental platform comprised a SeaBotix LBV 300 ROV with a Teledyne BlueView P900 FLS (130° FoV, 50 m max range, and 2.54 cm range resolution). To establish a ground-truth reference for localization tracking, a Differential GPS (DGPS) receiver and a magnetic heading compass were mounted on a surface-floating platform coupled directly to the vehicle. All benchmark evaluations were executed on a workstation with an Intel Core i9-14900K CPU, 64 GB of RAM, and an NVIDIA T1000 GPU.

\subsection{Feature Registration Performance}

\begin{tcolorbox}[
    colback=blue!5!white,     
    colframe=white!60!white,    
    arc=1mm,                   
    boxrule=0.1pt,             
    top=3mm, bottom=3mm,       
    left=3mm, right=3mm
]
\textbf{Research Question 1:} \textit{What is the performance impact of dense direct sonar registration compared to sparse keypoint methods?}

\smallskip
\textbf{Key Finding:} \textit{SOL-SLAM delivers the fastest optimization speeds and substantially reduces translation error—sustaining sub-meter precision across expanding displacements—while matching the rotational accuracy of sparse baselines.}
\end{tcolorbox}

\begin{tcolorbox}[
    colback=blue!5!white,     
    colframe=white!60!white,    
    arc=1mm,                   
    boxrule=0.1pt,             
    top=3mm, bottom=3mm,       
    left=3mm, right=3mm
]
\textbf{Research Question 2:} \textit{What is the performance impact of employing an Inverse Compositional Gauss-Newton (IC-GN) formulation relative to standard Forward Compositional Gradient Descent?}

\smallskip
\textbf{Key Finding:} \textit{The IC-GN approach yields superior tracking accuracy—demonstrating reduced translation and rotation errors—while simultaneously delivering drastically faster optimization speeds.}
\end{tcolorbox}

To evaluate the efficacy of dense sonar features against traditional sparse keypoint methods and to assess the impact of our optimization strategy, we analyze tracking precision under progressively increasing frame-to-frame displacement intervals. The dense direct registration approach is evaluated in two configurations: our proposed Inverse Compositional Gauss-Newton (IC-GN) formulation and a comparative Forward Compositional Gradient Descent (FC-GD) variant optimized with Adam. These are benchmarked against three prominent sparse feature extraction methods: AKAZE~\cite{AlcantarillaNB13} (a computationally efficient, handcrafted scale-space descriptor), SuperPoint~\cite{DeToneMR18} (a deep-learning-based approach optimized for visual imagery), and SONIC~\cite{GodeHK24} (a learned descriptor trained specifically on acoustic sonar data). For the sparse baselines, scan matching is executed via a Brute-Force (BF) matcher, and the rigid $\mathrm{SE}(2)$ transformation aligning successive scans is estimated using Random Sample Consensus (RANSAC).

Tracking robustness and convergence constraints are evaluated by pairing scans across frame-separation gaps of 5, 10, and 15 steps. This systematic increase in frame separation introduces larger initial misalignments, allowing us to characterize the capture range and stability of each method's convergence basin. The estimated transformation parameters (translations $t_x$, $t_y$ and heading angle $\theta$) are evaluated against the DGPS-compass ground truth using the root-mean-square error (RMSE) and mean computational runtime.

\begin{figure*}[ht]
\centering
\begin{subfigure}{0.32\linewidth}
\includegraphics[width=\textwidth]{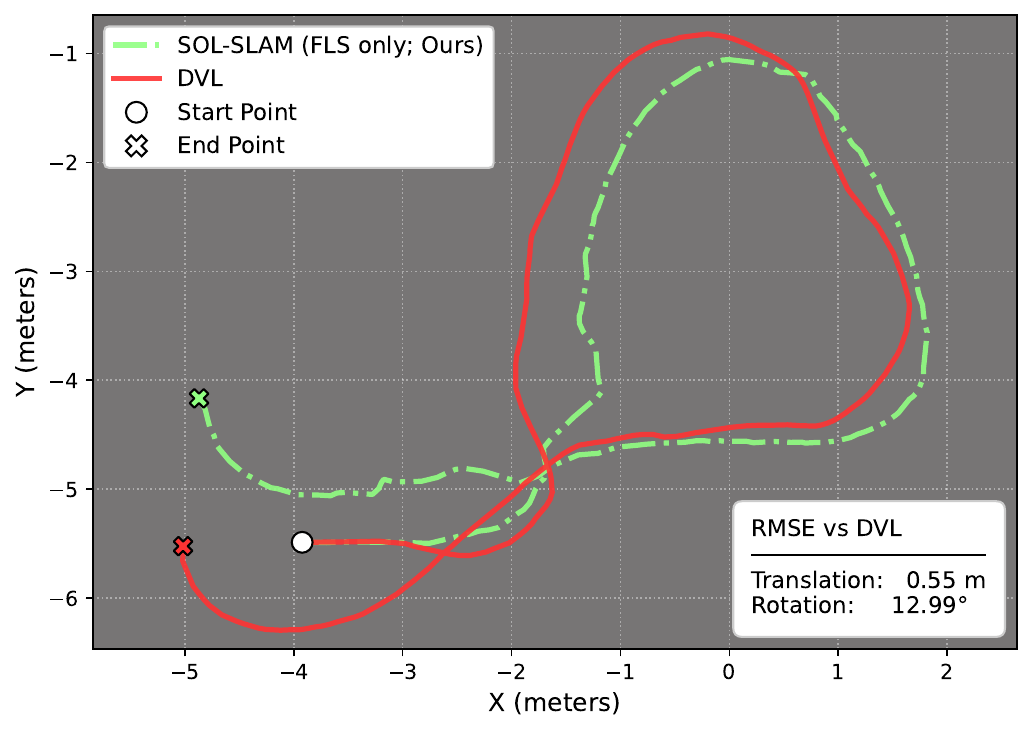}
\end{subfigure}
\hfill
\begin{subfigure}{0.32\linewidth}
\includegraphics[width=\textwidth]{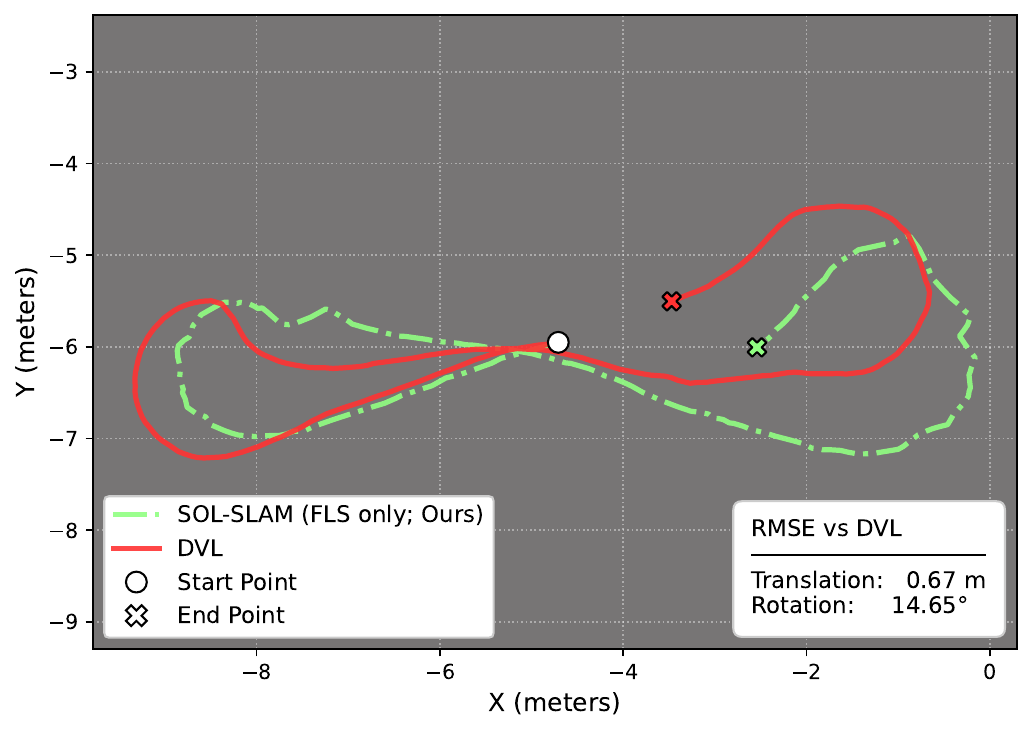}
\end{subfigure}
\hfill
\begin{subfigure}{0.32\linewidth}
\includegraphics[width=\textwidth]{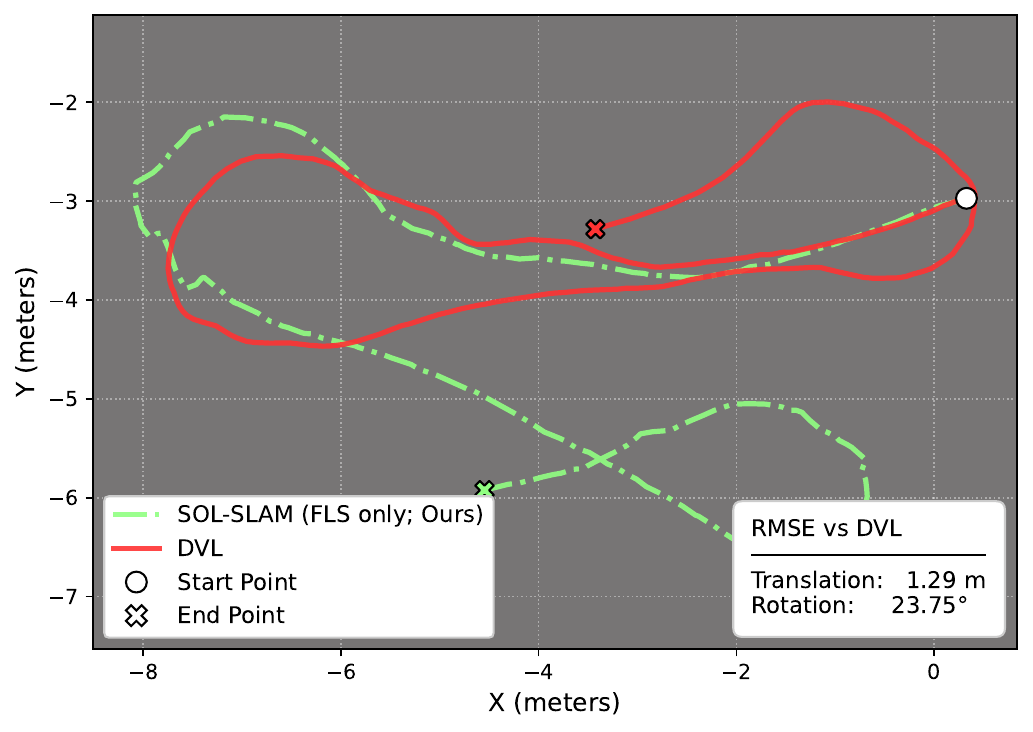}
\end{subfigure}
\caption{Estimated local trajectories from SOL-SLAM running fully onboard the Aqua2 AUV, evaluated against a DVL baseline. While accurate rotation estimation remains challenging, the results demonstrate the feasibility of onboard, sonar-only local SLAM on resource-constrained embedded hardware.}
\label{fig:aqua_trials}
\end{figure*}

Figure~\ref{fig:aracati_bar} demonstrates that the IC-GN SOL-SLAM approach substantially reduces translation error across all displacement thresholds, consistently maintaining sub-meter precision. Even under severe 15-frame gaps (representing a maximum translation displacement of $1.48~\text{m}$ and angular displacement of $34.14^\circ$), our dense direct approach demonstrates exceptional robustness. By effectively leveraging the full acoustic intensity profile, SOL-SLAM sustains tight alignment where sparse baselines (AKAZE, SuperPoint, and SONIC) yield translation errors ranging from $7.5~\text{m}$ to over $24~\text{m}$.

When evaluating the two dense optimization strategies, the IC-GN formulation unlocks significant performance gains over the FC-GD approach. As shown in Figure~\ref{fig:aracati_bar}, IC-GN achieves tighter bounds on both translation and rotation errors across all tested separation intervals compared to the Adam-optimized FC-GD variant. Most notably, the IC-GN strategy efficiently precomputes spatial gradients to drastically accelerate inference. By eliminating the computational overhead of continuous gradient re-evaluation, the IC-GN approach achieves an execution latency of approximately $40~\text{ms}$ ($25~\text{Hz}$), well within the real-time processing constraints required for closed-loop subsea navigation. While rotational error for the IC-GN approach scales slightly at wider frame separations, this variation is easily mitigated in practical deployments by bounding heading drift with a low-cost Inertial Measurement Unit (IMU)—a standard sensor payload on autonomous underwater platforms.

\subsection{Comparison with Multi-Sensor Odometry Approaches}

\begin{tcolorbox}[
    colback=blue!5!white,     
    colframe=white!60!white,    
    arc=1mm,                   
    boxrule=0.1pt,             
    top=3mm, bottom=3mm,       
    left=3mm, right=3mm
]
\textbf{Research Question 3:} \textit{How does the proposed FLS-only local SLAM approach compare against multi-sensor systems that fuse FLS, DVL, and IMU data with loop closure?}

\smallskip
\textbf{Key Finding:} \textit{The standalone, FLS-only SOL-SLAM approach achieves highly competitive trajectory estimation accuracy relative to a state-of-the-art multi-sensor odometry system, despite operating entirely without auxiliary velocity or inertial data.}
\end{tcolorbox}

We evaluate the capability of the proposed SOL-SLAM approach to sustain continuous local trajectory estimation in the complete absence of auxiliary velocity or inertial measurements. SOL-SLAM is benchmarked against DISO~\cite{XuZHLW24}, a state-of-the-art multi-sensor underwater odometry system that ensures robust tracking by tightly coupling FLS registration with synchronous Doppler Velocity Log (DVL) and IMU inputs. Benchmarking is performed across two dataset subsequences where sufficient acoustic feature density is preserved, with each active frame retaining a minimum of 1\% sonar features above the operational intensity threshold.

Because the physical platform deployed for the Aracati dataset lacked a physical DVL, the DISO baseline utilizes the vehicle's internal commanded control velocities as a proxy for dead-reckoning sensor feedback. Since the vehicle relied on high-accuracy DGPS for closed-loop navigation, it closely tracked these commanded trajectories, rendering the control inputs a reliable substitute for true DVL measurements within the baseline system.

The estimated trajectories are illustrated in Figure~\ref{fig:aracati_trace}. In Subsequence 1, DISO yields a translation RMSE of $0.57~\text{m}$ and a rotation RMSE of $2.85^\circ$, whereas SOL-SLAM achieves a competitive translation RMSE of $1.76~\text{m}$ and a rotation RMSE of $3.16^\circ$. In Subsequence 2, the performance gap narrows significantly: our approach tracks the trajectory with a translation RMSE of $0.92~\text{m}$ and a rotation RMSE of $4.30^\circ$, closely matching the performance of DISO ($0.65~\text{m}$ and $2.63^\circ$). These benchmarks validate that our FLS-only SOL-SLAM approach achieves local SLAM precision comparable to multi-sensor configurations, eliminating the need for expensive auxiliary sensors in feature rich environments.

\subsection{AUV Field Trial with Onboard Processing}

\begin{tcolorbox}[
    colback=blue!5!white,     
    colframe=white!60!white,    
    arc=1mm,                   
    boxrule=0.1pt,             
    top=3mm, bottom=3mm,       
    left=3mm, right=3mm
]
\textbf{Research Question 4:} \textit{Can SOL-SLAM sustain fast local SLAM onboard an AUV?}

\smallskip
\textbf{Key Finding:} \textit{Executing onboard an NVIDIA Jetson Orin Nano SoC at $8\text{ Hz}$, SOL-SLAM demonstrates the feasibility of fully self-contained local SLAM.}
\end{tcolorbox}

We validated SOL-SLAM in real-world field trials using an Aqua2 AUV~\cite{DudekGPSSTJGHRZMLZBG07} (Fig.~\ref{fig:aqua}) equipped with a Sonoptix Echo forward-looking sonar (120° FoV, 12~m max range, and 8 mm range resolution). The vehicle was manually piloted in depth-hold mode inside a swimming pool, while SOL-SLAM executed fully onboard an embedded NVIDIA Jetson Nano SoC. Figure~\ref{fig:aqua_trials} compares the trajectory estimated by SOL-SLAM with DVL-based odometry used as a proxy ground truth.

Operating at $8\text{ Hz}$, SOL-SLAM achieves sufficient throughput for closed-loop navigation on resource-constrained embedded hardware. Consistent with benchmark evaluations, while rotational drift remains a challenge, the system accurately resolves translational motion and closely tracks the DVL baseline.
\section{Limitations and Future Work}

\textbf{Rotational Sensitivity:} SOL-SLAM currently struggles with rapid rotational motion. Future iterations could address this by fusing data from an on-board IMU---a low-cost sensor present on most AUV platforms.

\textbf{Scan-to-Map Drift:} To prevent substantial drift from occasional scan-matching failures, future work will explore leveraging IMU measurements for mismatch detection and drift suppression without the computational overhead of full pose-graph loop closure.

\textbf{Feature Sparsity:} Like most SLAM approaches, performance degrades in acoustically degraded or featureless environments. While the extended sensing range of forward-looking sonar (often exceeding $100\text{ m}$) helps maintain trackable returns, future work will integrate SOL-SLAM with active, perception-aware navigation strategies (e.g., AquaVis~\cite{XanthidisKKJVOR21}) to autonomously guide the vehicle toward feature-rich regions.
\section{Conclusion}
This paper presented SOL-SLAM, a direct local SLAM approach operating exclusively on Forward-Looking Sonar (FLS) imagery. While conventional methods rely on sparse feature extraction—often discarding valuable information in already information-sparse acoustic returns—our method directly leverages full, dense acoustic scans. To mitigate the computational overhead of dense alignment, we introduced an efficient inverse compositional direct registration approach that achieves real-time performance.

Key contributions and findings include:
\begin{itemize}
    \item \textbf{Direct Registration Accuracy:} SOL-SLAM achieves sub-meter translational accuracy over large displacements, consistently outperforming sparse keypoint methods (AKAZE, SuperPoint, and SONIC) vulnerable to acoustic mismatching.
    \item \textbf{Competitive with Multi-Sensor Odometry:} Operating without auxiliary inertial (IMU) or Doppler velocity (DVL) measurements, our FLS-only approach achieves accuracy competitive with state-of-the-art multi-modal odometry (DISO).
    \item \textbf{Field Trial Validation:} Field trials with an FLS-equipped AUV verified the method's real-world viability, demonstrating reliable state estimation executing entirely on the vehicle's onboard embedded computer.
\end{itemize}

\bibliography{references}

\end{document}